\def\year{2020}\relax
\documentclass[letterpaper]{article} 
\usepackage{aaai20}  
\usepackage{times}  
\usepackage{helvet} 
\usepackage{courier}  
\usepackage[hyphens]{url}  
\usepackage{graphicx} 
\usepackage{graphicx}  
\usepackage{amsmath,amssymb,amsthm,bbm,mathtools,bm}
\usepackage{algorithm,algorithmicx,algpseudocode}
\usepackage{subfigure}
\usepackage{color}

\graphicspath{{../docs/figs/}}

\newcommand{\real}{\mathbb{R}}

\DeclareMathOperator{\diag}{diag}

\DeclareMathOperator{\sigmoid}{sigmoid}

\DeclareMathOperator{\gru}{GRU}
\DeclareMathOperator{\lstm}{LSTM}
\DeclareMathOperator{\gconv}{GCONV}

\DeclareMathOperator{\egcuh}{EGCU-H}
\DeclareMathOperator{\egcuo}{EGCU-O}

\title{EvolveGCN: Evolving Graph Convolutional Networks for Dynamic Graphs}
\author{
Aldo Pareja,$^{1,2}$\thanks{Equal contribution}
Giacomo Domeniconi,$^{1,2}$\footnotemark[1]
Jie Chen,$^{1,2}$\thanks{Contact author}
Tengfei Ma,$^{1,2}$
Toyotaro Suzumura,$^{1,2}$\\
\Large \textbf{%
Hiroki Kanezashi,$^{1,2}$
Tim Kaler,$^{1,3}$
Tao B. Schardl,$^{1,3}$
Charles E. Leiserson$^{1,3}$}\\
$^1$MIT-IBM Watson AI Lab,
$^2$IBM Research,
$^3$MIT CSAIL\\
\{Aldo.Pareja, Giacomo.Domeniconi1\}@ibm.com, chenjie@us.ibm.com\\
Tengfei.Ma1@ibm.com, \{tsuzumura, hirokik\}@us.ibm.com,
\{tfk, neboat, cel\}@mit.edu
}
 
\begin{document}

\maketitle

\begin{abstract}
  Graph representation learning resurges as a trending research subject owing to the widespread use of deep learning for Euclidean data, which inspire various creative designs of neural networks in the non-Euclidean domain, particularly graphs. With the success of these graph neural networks (GNN) in the static setting, we approach further practical scenarios where the graph dynamically evolves. Existing approaches typically resort to node embeddings and use a recurrent neural network (RNN, broadly speaking) to regulate the embeddings and learn the temporal dynamics. These methods require the knowledge of a node in the full time span (including both training and testing) and are less applicable to the frequent change of the node set. In some extreme scenarios, the node sets at different time steps may completely differ. To resolve this challenge, we propose EvolveGCN, which adapts the graph convolutional network (GCN) model along the temporal dimension without resorting to node embeddings. The proposed approach captures the dynamism of the graph sequence through using an RNN to evolve the GCN parameters. Two architectures are considered for the parameter evolution. We evaluate the proposed approach on tasks including link prediction, edge classification, and node classification. The experimental results indicate a generally higher performance of EvolveGCN compared with related approaches. The code is available at \url{https://github.com/IBM/EvolveGCN}.
\end{abstract}

\section{Introduction}
Graphs are ubiquitous data structures that model the pairwise interactions between entities. Learning with graphs encounters unique challenges, including their combinatorial nature and the scalability bottleneck, compared with Euclidean data (e.g., images, videos, speech signals, and natural languages). With the remarkable success of deep learning for the latter data types, there exist renewed interests in the learning of graph representations~\cite{Perozzi2014,Tang2015,Cao2015,Ou2016,Grover2016} on both the node and the graph level, now parameterized by deep neural networks~\cite{Bruna2014,Duvenaud2015,Defferrard2016,Li2016,Gilmer2017,Kipf2017,Hamilton2017,Jin2017,Chen2018,Velickovic2018,Gao2019}.

These neural network models generally focus on a given, static graph. In real-life applications, however, often one encounters a dynamically evolving graph. For example, users of a social network develop friendship over time; hence, the vectorial representation of the users should be updated accordingly to reflect the temporal evolution of their social relationship. Similarly, a citation network of scientific articles is constantly enriched due to frequent publications of new work citing prior art. Thus, the influence, and even sometimes the categorization, of an article varies along time. Update of the node embeddings to reflect this variation is desired. In financial networks, transactions naturally come with time stamps. The nature of a user account may change owing to the characteristics of the involved transactions (e.g., an account participates money laundering or a user becomes a victim of credit card fraud). Early detection of the change is crucial to the effectiveness of law enforcement and the minimization of loss to a financial institute. These examples urge the development of dynamic graph methods that encode the temporal evolution of relational data.

Built on the recent success of graph neural networks (GNN) for static graphs, in this work we extend them to the dynamic setting through introducing a recurrent mechanism to update the network parameters, for capturing the dynamism of the graphs. A plethora of GNNs perform information fusion through aggregating node embeddings from one-hop neighborhoods recursively. A majority of the parameters of the networks is the linear transformation of the node embeddings in each layer. We specifically focus on the graph convolutional network (GCN)~\cite{Kipf2017} because of its simplicity and effectiveness. Then, we propose to use a recurrent neural network (RNN) to inject the dynamism into the parameters of the GCN, which forms an evolving sequence.

Work along a similar direction includes~\cite{Seo2016,Manessia2017,Narayan2018}, among others, which are based on a combination of GNNs (typically GCN) and RNNs (typically LSTM). These methods use GNNs as a feature extractor and RNNs for sequence learning from the extracted features (node embeddings). As a result, one single GNN model is learned for all graphs on the temporal axis. A limitation of these methods is that they require the knowledge of the nodes over the whole time span and can hardly promise the performance on new nodes in the future.

In practice, in addition to the likelihood that new nodes may emerge after training, nodes may also frequently appear and disappear, which renders the node embedding approaches questionable, because it is challenging for RNNs to learn these irregular behaviors. To resolve these challenges, we propose instead to use the RNN to regulate the GCN model (i.e., network parameters) at every time step. This approach effectively performs model adaptation, which focuses on the model itself rather than the node embeddings. Hence, change of nodes poses no restriction. Further, for future graphs with new nodes without historical information, the evolved GCN is still sensible for them.

Note that in the proposed method, the GCN parameters are not trained anymore. They are computed from the RNN and hence only the RNN parameters are trained. In this manner, the number of parameters (model size) does not grow with the number of time steps and the model is as manageable as a typical RNN.

\section{Related Work}
Methods for dynamic graphs are often extensions of those for a static one, with an additional focus on the temporal dimension and update schemes. For example, in matrix factorization-based approaches~\cite{Roweis2000,Belkin2002}, node embeddings come from the (generalized) eigenvectors of the graph Laplacian matrix. Hence, DANE~\cite{Li2017} updates the eigenvectors efficiently based on the prior ones, rather than computing them from scratch for each new graph. The dominant advantage of such methods is the computational efficiency.

For random walk-based approaches~\cite{Perozzi2014,Grover2016}, transition probabilities conditioned on history are modeled as the normalized inner products of the corresponding node embeddings. These approaches maximize the probabilities of the sampled random walks. CTDANE~\cite{Nguyen2018} extends this idea by requiring the walks to obey the temporal order. Another work, NetWalk~\cite{Yu2018a}, does not use the probability as the objective function; rather, it observes that if the graph does not undergo substantial changes, one only needs to resample a few walks in the successive time step. Hence, this approach incrementally retrains the model with warm starts, substantially reducing the computational cost.

The wave of deep learning introduces a flourish of unsupervised and supervised approaches for parameterizing the quantities of interest with neural networks. DynGEM~\cite{Goyal2017} is an autoencoding approach that minimizes the reconstruction loss, together with the distance between connected nodes in the embedding space. A feature of DynGEM is that the depth of the architecture is adaptive to the size of the graph; and the autoencoder learned from the past time step is used to initialize the training of the one in the following time.

A popular category of approaches for dynamic graphs is point processes that are continuous in time. Know-Evolve~\cite{Trivedi2017} and DyRep~\cite{Trivedi2018} model the occurrence of an edge as a point process and parameterize the intensity function by using a neural network, taking node embeddings as the input. DynamicTriad~\cite{Zhou2018} uses a point process to model a more complex phenomenon---triadic closure---where a triad with three nodes is developed from an open one (a pair of nodes are not connected) to a closed one (all three pairs are connected). HTNE~\cite{Zuo2018} similarly models the dynamism by using the Hawkes process, with additionally an attention mechanism to determine the influence of historical neighbors on the current neighbors of a node. These methods are advantageous for event time prediction because of the continuous nature of the process.

A set of approaches most relevant to this work is combinations of GNNs and recurrent architectures (e.g., LSTM), whereby the former digest graph information and the latter handle dynamism. The most explored GNNs in this context are of the convolutional style and we call them graph convolutional networks (GCN), following the terminology of the related work, although in other settings GCN specifically refers to the architecture proposed by~\cite{Kipf2017}. GCRN~\cite{Seo2016} offers two combinations. The first one uses a GCN to obtain node embeddings, which are then fed into the LSTM that learns the dynamism. The second one is a modified LSTM that takes node features as input but replaces the fully connected layers therein by graph convolutions. The first idea is similarly explored in WD-GCN/CD-GCN~\cite{Manessia2017} and RgCNN~\cite{Narayan2018}. WD-GCN/CD-GCN modifies the graph convolution layers, most notably by adding a skip connection. In addition to such simple combinations, STGCN~\cite{Yu2018} proposes a complex architecture that consists of so-called ST-Conv blocks. In this model, the node features must be evolving over time, since inside each ST-Conv block, a 1D convolution of the node features is first performed along the temporal dimension, followed by a graph convolution and another 1D convolution. This architecture was demonstrated for spatiotemporal traffic data (hence the names STGCN and ST-Conv), where the spatial information is handled by using graph convolutions.

\begin{figure*}[ht]
  \centering
  \includegraphics[width=.8\linewidth]{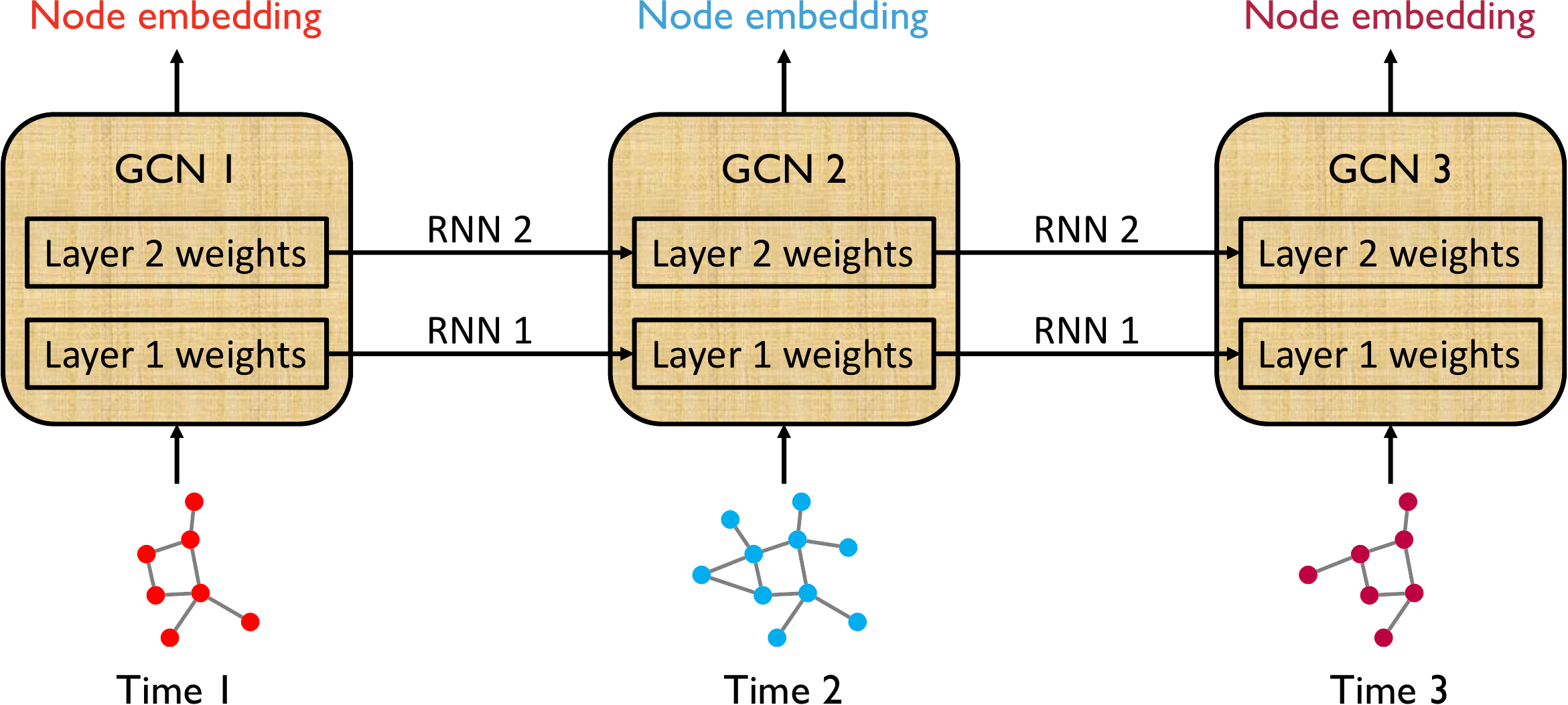}
  \caption{Schematic illustration of EvolveGCN. The RNN means a recurrent architecture in general (e.g., GRU, LSTM). We suggest two options to evolve the GCN weights, treating them with different roles in the RNN. See the EvolveGCN-H version and EvolveGCN-O version in Figure~\ref{fig:egcu}.} 
  \label{fig:egcn}
\end{figure*}

\section{Method}
In this section we present a novel method, coined \emph{evolving graph convolutional network} (EvolveGCN), that captures the dynamism underlying a graph sequence by using a recurrent model to evolve the GCN parameters. Throughout we will use subscript $t$ to denote the time index and superscript $l$ to denote the GCN layer index. To avoid notational cluttering, we assume that all graphs have $n$ nodes; although we reiterate that the node sets, as well as the cardinality, may change over time. Then, at time step $t$, the input data consists of the pair $(A_t\in\real^{n\times n}, X_t\in\real^{n\times d})$, where the former is the graph (weighted) adjacency matrix and the latter is the matrix of input node features. Specifically, each row of $X_t$ is a $d$-dimensional feature vector of the corresponding node.

\subsection{Graph Convolutional Network (GCN)}\label{sec:gconv}
A GCN~\cite{Kipf2017} consists of multiple layers of graph convolution, which is similar to a perceptron but additionally has a neighborhood aggregation step motivated by spectral convolution. At time $t$, the $l$-th layer takes the adjacency matrix $A_t$ and the node embedding matrix $H_t^{(l)}$ as input, and uses a weight matrix $W_t^{(l)}$ to update the node embedding matrix to $H_t^{(l+1)}$ as output. Mathematically, we write
\begin{align}
  H_t^{(l+1)} &= \gconv(A_t, H_t^{(l)}, W_t^{(l)}) \nonumber \\
  &:= \sigma(\widehat{A}_t H_t^{(l)} W_t^{(l)}), \label{eqn:gcn}
\end{align}
where $\widehat{A}_t$ is a normalization of $A_t$ defined as (omitting time index for clarity):
\[
\widehat{A}=\widetilde{D}^{-\frac{1}{2}}\widetilde{A}\widetilde{D}^{-\frac{1}{2}},
\quad \widetilde{A}=A+I,
\quad \widetilde{D}=\diag\Bigg(\sum_j\widetilde{A}_{ij}\Bigg),
\]
and $\sigma$ is the activation function (typically ReLU) for all but the output layer. The initial embedding matrix comes from the node features; i.e., $H_t^{(0)}=X_t$. Let there be $L$ layers of graph convolutions. For the output layer, the function $\sigma$ may be considered the identity, in which case $H_t^{(L)}$ contains high-level representations of the graph nodes transformed from the initial features; or it may be the softmax for node classification, in which case $H_t^{(L)}$ consists of prediction probabilities.

\begin{figure*}[ht]
  \centering
  \subfigure[EvolveGCN-H, where the GCN parameters are hidden states of a recurrent architecture that takes node embeddings as input.]{\includegraphics[width=.9\linewidth]{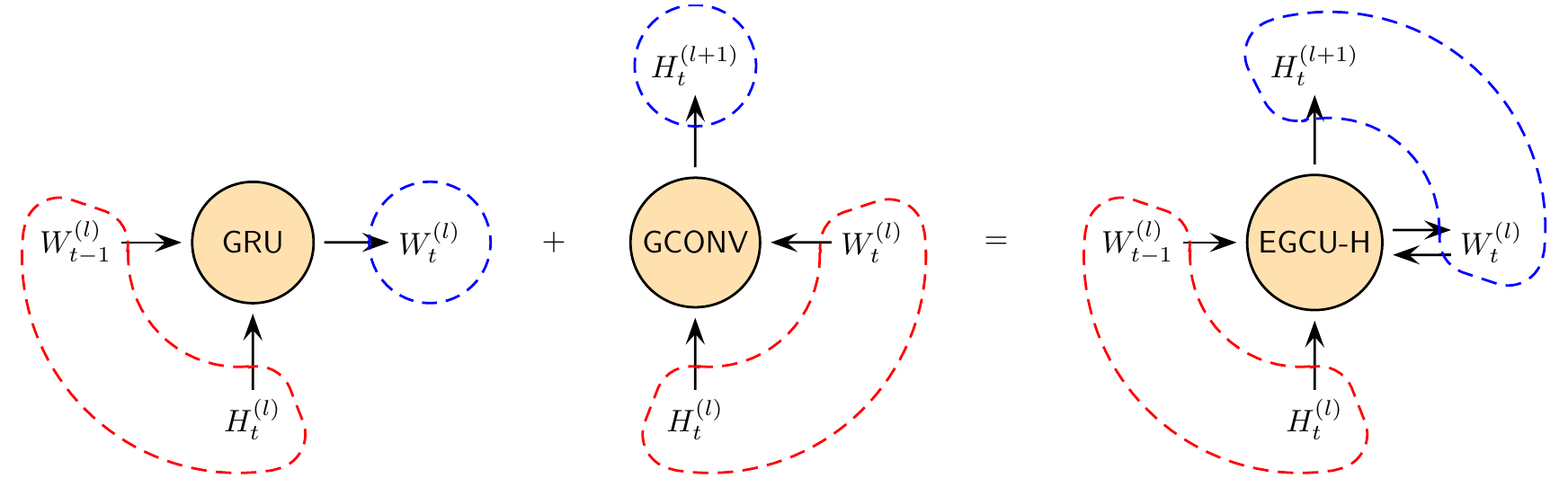}}
  \subfigure[EvolveGCN-O, where the GCN parameters are input/outputs of a recurrent architecture.]{\includegraphics[width=.9\linewidth]{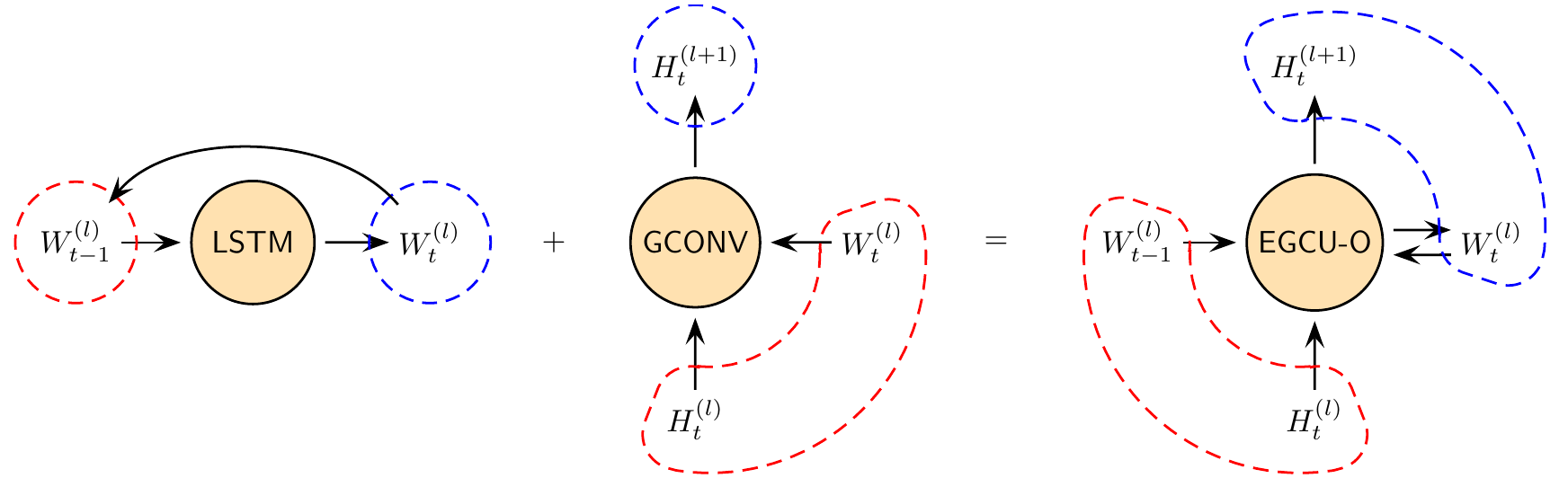}}
  \caption{Two versions of EvolveGCN. In each version, the left is a recurrent architecture; the middle is the graph convolution unit; and the right is the evolving graph convolution unit. Red region denotes information input to the unit and blue region denotes output information. The mathematical notation $W$ means GCN parameters and $H$ means node embeddings. Time $t$ progresses from left to right, whereas neural network layers $l$ are built up from bottom to top.}
  \label{fig:egcu}
\end{figure*}

Figure~\ref{fig:egcn} is a schematic illustration of the proposed EvolveGCN, wherein each time step contains one GCN indexed by time. The parameters of the GCN are the weight matrices $W_t^{(l)}$, for different time steps $t$ and layers $l$. Graph convolutions occur for a particular time but generate new information along the layers. Figure~\ref{fig:egcu} illustrates the computation at each layer. The relationship between $H_t^{(l)}$, $W_t^{(l)}$, and $H_t^{(l+1)}$ is depicted in the middle part of the figure.

\subsection{Weight Evolution}\label{sec:gru}
At the heart of the proposed method is the update of the weight matrix $W_t^{(l)}$ at time $t$ based on current, as well as historical, information. This requirement can be naturally fulfilled by using a recurrent architecture, with two options.

The first option is to treat $W_t^{(l)}$ as the hidden state of the dynamical system. We use a gated recurrent unit (GRU) to update the hidden state upon time-$t$ input to the system. The input information naturally is the node embeddings $H_t^{(l)}$. Abstractly, we write
\[
\overbrace{\underbrace{W_t^{(l)}}_{\text{hidden state}}}^{\text{GCN weights}}
= \gru(
\overbrace{\underbrace{H_t^{(l)}}_{\text{input}}}^{\text{node embeddings}},
\overbrace{\underbrace{W_{t-1}^{(l)}}_{\text{hidden state}}}^{\text{GCN weights}}
),
\]
with details deferred to a later subsection. The GRU may be replaced by other recurrent architectures, as long as the roles of $W_t^{(l)}$, $H_t^{(l)}$, and $W_{t-1}^{(l)}$ are clear. We use ``-H'' to denote this version; see the left part of Figure~\ref{fig:egcu}(a).

The second option is to treat $W_t^{(l)}$ as the output of the dynamical system (which becomes the input at the subsequent time step). We use a long short-term memory (LSTM) cell to model this input-output relationship. The LSTM itself maintains the system information by using a cell context, which acts like the hidden state of a GRU. In this version, node embeddings are not used at all. Abstractly, we write
\[
\overbrace{\underbrace{W_t^{(l)}}_{\text{output}}}^{\text{GCN weights}}
= \lstm(
\overbrace{\underbrace{W_{t-1}^{(l)}}_{\text{input}}}^{\text{GCN weights}}
),
\]
with details deferred to a later subsection. The LSTM may be replaced by other recurrent architectures, as long as the roles of $W_t^{(l)}$ and $W_{t-1}^{(l)}$ are clear. We use ``-O'' to denote this version; see the left part of Figure~\ref{fig:egcu}(b).

\subsection{Evolving Graph Convolution Unit (EGCU)}
Combining the graph convolution unit $\gconv$ presented in Section~\ref{sec:gconv} and a recurrent architecture presented in Section~\ref{sec:gru}, we reach the \emph{evolving graph convolution unit} (EGCU). Depending on the way that GCN weights are evolved, we have two versions:
\begin{algorithmic}[1]
  \baselineskip=15pt
  \Function {$[H_t^{(l+1)}, W_t^{(l)}] = \egcuh$}{$A_t, H_t^{(l)}, W_{t-1}^{(l)}$}
  \State $W_t^{(l)} = \gru(H_t^{(l)}, W_{t-1}^{(l)})$
  \State $H_t^{(l+1)} = \gconv(A_t, H_t^{(l)}, W_t^{(l)})$
  \EndFunction
\end{algorithmic}
\begin{algorithmic}[1]
  \baselineskip=15pt
  \Function {$[H_t^{(l+1)}, W_t^{(l)}] = \egcuo$}{$A_t, H_t^{(l)}, W_{t-1}^{(l)}$}
  \State $W_t^{(l)} = \lstm(W_{t-1}^{(l)})$
  \State $H_t^{(l+1)} = \gconv(A_t, H_t^{(l)}, W_t^{(l)})$
  \EndFunction
\end{algorithmic}
In the -H version, the GCN weights are treated as hidden states of the recurrent architecture; whereas in the -O version, these weights are treated as input/outputs. In both versions, the EGCU performs graph convolutions along layers and meanwhile evolves the weight matrices over time.

Chaining the units bottom-up, we obtain a GCN with multiple layers for one time step. Then, unrolling over time horizontally, the units form a lattice on which information ($H_t^{(l)}$ and $W_t^{(l)}$) flows. We call the overall model \emph{evolving graph convolutional network} (EvolveGCN).

\subsection{Implementation of the -H Version}
The -H version can be implemented by using a standard GRU, with two extensions: (a) extending the inputs and hidden states from vectors to matrices (because the hidden state is now the GCN weight matrices); and (b) matching the column dimension of the input with that of the hidden state.

The matrix extension is straightforward: One simply places the column vectors side by side to form a matrix. In other words, one uses the same GRU to process each column of the GCN weight matrix. For completeness, we write the matrix version of GRU in the following, by noting that all named variables (such as $X_t$ and $H_t$) are only local variables; they are not to be confused with the mathematical notations we have been using so far. We use these local variable names so that the reader easily recognizes the GRU functionality.

\begin{algorithmic}[1]
  \baselineskip=15pt
  \Function {$H_t=g$}{$X_t, H_{t-1}$}
  \State $Z_t = \sigmoid(W_ZX_t + U_ZH_{t-1} + B_Z)$
  \State $R_t = \sigmoid(W_RX_t + U_RH_{t-1} + B_R)$
  \State $\widetilde{H}_t = \tanh(W_HX_t + U_H(R_t\circ H_{t-1})+B_H)$
  \State $H_t = (1-Z_t) \circ H_{t-1} + Z_t \circ \widetilde{H}_t$
  \EndFunction
\end{algorithmic}

The second requirement is that the number of columns of the GRU input must match that of the hidden state. Let the latter number be $k$. Our strategy is to summarize all the node embedding vectors into $k$ representative ones (each used as a column vector). The following pseudocode gives one popular approach for this summarization. By convention, it takes a matrix $X_t$ with many rows as input and produces a matrix $Z_t$ with only $k$ rows (see, e.g., \cite{Cangea2018,Gao2019}). The summarization requires a parameter vector $p$ that is independent of the time index $t$ (but may vary for different graph convolution layers). This vector is used to compute weights for the rows, among which the ones corresponding to the top $k$ weights are selected and are weighted for output.

\begin{algorithmic}[1]
  \baselineskip=15pt
  \Function {$Z_t=summarize$}{$X_t,k$}
  \State $y_t = X_tp/\|p\|$
  \State $i_t = \text{top-indices}(y_t, k)$
  \State $Z_t = [X_t \circ \tanh(y_t)]_{i_t}$
  \EndFunction
\end{algorithmic}

With the above functions $g$ and $summarize$, we now completely specify the recurrent architecture:
\begin{align*}
  W_t^{(l)} &= \gru(H_t^{(l)}, W_{t-1}^{(l)}) \\
  &:= g(summarize(H_t^{(l)}, \#col(W_{t-1}^{(l)}))^T, W_{t-1}^{(l)}),
\end{align*}
where $\#col$ denotes the number of columns of a matrix and the superscript $T$ denotes matrix transpose. Effectively, it summarizes the node embedding matrix $H_t^{(l)}$ into one with appropriate dimensions and then evolves the weight matrix $W_{t-1}^{(l)}$ in the past time step to $W_t^{(l)}$ for the current time.

Note again that the recurrent hidden state may be realized by not only GRU, but also other RNN architectures as well.

\subsection{Implementation of the -O Version}
Implementing the -O version requires only a straightforward extension of the standard LSTM from the vector version to the matrix version. The following is the pseudocode, where note again that all named variables are only local variables and they are not to be confused with the mathematical notations we have been using so far. We use these local variable names so that the reader easily recognizes the LSTM functionality.

\begin{algorithmic}[1]
  \baselineskip=15pt
  \Function {$H_t=f$}{$X_t$}
  \State Current input $X_t$ is the same as the past output $H_{t-1}$
  \State $F_t = \sigmoid(W_FX_t + U_FH_{t-1} + B_F)$
  \State $I_t = \sigmoid(W_IX_t + U_IH_{t-1} + B_I)$
  \State $O_t = \sigmoid(W_OX_t + U_OH_{t-1} + B_O)$
  \State $\widetilde{C}_t = \tanh(W_CX_t + U_CH_{t-1}+B_C)$
  \State $C_t = F_t \circ C_{t-1} + I_t \circ \widetilde{C}_t$
  \State $H_t = O_t \circ \tanh(C_t)$
  \EndFunction
\end{algorithmic}

With the above function $f$, we now completely specify the recurrent architecture:
\[
  W_t^{(l)} = \lstm(W_{t-1}^{(l)})
  := f(W_{t-1}^{(l)}).
\]
Note again that the recurrent input-output relationship may be realized by not only LSTM, but also other RNN architectures as well.

\subsection{Which Version to Use}
Choosing the right version is data set dependent. When node features are informative, the -H version may be more effective, because it incorporates additionally node embedding in the recurrent network. On the other hand, if the node features are not much informative but the graph structure plays a more vital role, the -O version focuses on the change of the structure and may be more effective.

\section{Experiments}
In this section, we present a comprehensive set of experiments to demonstrate the effectiveness of EvolveGCN. The setting includes a variety of data sets, tasks, compared methods, and evaluation metrics. Hyperparameters are tuned by using the validation set and test results are reported at the best validation epoch.

\subsection{Data Sets}
We use a combination of synthetic and publicly available benchmark data sets for experiments.

\vskip5pt \noindent
\textbf{Stochastic Block Model.}
(SBM for short)
SBM is a popularly used random graph model for simulating community structures and evolutions. We follow~\cite{Goyal2017} to generate synthetic data from the model.

\vskip5pt \noindent
\textbf{Bitcoin OTC.}%
\footnote{\url{http://snap.stanford.edu/data/soc-sign-bitcoin-otc.html}}
(BC-OTC for short)
BC-OTC is a who-trusts-whom network of bitcoin users trading on the platform \url{http://www.bitcoin-otc.com}. The data set may be used for predicting the polarity of each rating and forecasting whether a user will rate another one in the next time step.

\vskip5pt \noindent
\textbf{Bitcoin Alpha.}%
\footnote{\url{http://snap.stanford.edu/data/soc-sign-bitcoin-alpha.html}}
(BC-Alpha for short)
BC-Alpha is created in the same manner as is BC-OTC, except that the users and ratings come from a different trading platform, \url{http://www.btc-alpha.com}.

\vskip5pt \noindent
\textbf{UC Irvine messages.}%
\footnote{\url{http://konect.uni-koblenz.de/networks/opsahl-ucsocial}}
(UCI for short)
UCI is an online community of students from the University of California, Irvine, wherein the links of this social network indicate sent messages between users. Link prediction is a standard task for this data set.

\vskip5pt \noindent
\textbf{Autonomous systems.}%
\footnote{\url{http://snap.stanford.edu/data/as-733.html}}
(AS for short)
AS is a communication network of routers that exchange traffic flows with peers. This data set may be used to forecast message exchanges in the future.

\vskip5pt \noindent
\textbf{Reddit Hyperlink Network.}%
\footnote{\url{http://snap.stanford.edu/data/soc-RedditHyperlinks.html}}
(Reddit for short)
Reddit is a subreddit-to-subreddit hyperlink network, where each hyperlink originates from a post in the source community and links to a post in the target community. The hyperlinks are annotated with sentiment. The data set may be used for sentiment classification.

\vskip5pt \noindent
\textbf{Elliptic.}%
\footnote{\url{https://www.kaggle.com/ellipticco/elliptic-data-set}}
Elliptic is a network of bitcoin transactions, wherein each node represents one transaction and the edges indicate payment flows. Approximately 20\% of the transactions have been mapped to real entities belonging to licit categories versus illicit ones. The aim is to categorize the unlabeled transactions.

These data sets are summarized in Table~\ref{tab:dataset}. Training/validation/test splits are done along the temporal dimension. The temporal granularity is case dependent but we use all available information of the data sets, except AS for which we use only the first 100 days following~\cite{Goyal2017}.

\begin{table}[ht]
\centering
\caption{Data sets.}
\label{tab:dataset}
\begin{tabular}{cccc}
\hline
& \# Nodes & \# Edges & \# Time Steps\\
&          &          & (Train / Val / Test) \\
\hline
SBM      & 1,000   & 4,870,863 & 35 / 5 / 10 \\
BC-OTC   & 5,881   & 35,588  & 95 / 14 / 28 \\
BC-Alpha & 3,777   & 24,173  & 95 / 13 / 28 \\
UCI      & 1,899   & 59,835  & 62 / 9 / 17 \\
AS       & 6,474   & 13,895  & 70 / 10 / 20 \\ 
Reddit   & 55,863  & 858,490 & 122 / 18 / 34 \\
Elliptic & 203,769 & 234,355 & 31 / 5 / 13 \\
\hline
\end{tabular}
\end{table}

\subsection{Tasks}
The proposed EvolveGCN supports three predictive tasks elaborated below. The model for producing the embeddings and the predictive model are trained end to end. The output embedding of a node $u$ by GCN at time $t$ is denoted by $h_t^u$.

\vskip5pt \noindent
\textbf{Link Prediction.}
The task of link prediction is to leverage information up to time $t$ and predict the existence of an edge $(u,v)$ at time $t+1$. Since historical information has been encoded in the GCN parameters, we base the prediction on $h_t^u$ and $h_t^v$. To achieve so, we concatenate these two vectors and apply an MLP to obtain the link probability. As a standard practice, we perform negative sampling and optimize the cross-entropy loss function.

Five data sets are used for experimentation for this task. See the header of Table~\ref{tab:link_prediction}. Evaluation metrics include mean average precision (MAP) and mean reciprocal rank (MRR).

\vskip5pt \noindent
\textbf{Edge Classification.}
Predicting the label of an edge $(u,v)$ at time $t$ is done in almost the same manner as link prediction: We concatenate $h_t^u$ and $h_t^v$ and apply an MLP to obtain the class probability. 

Three data sets are used for experimentation for this task: BC-OTC, BC-Alpha, and Reddit. Evaluation metrics are precision, recall, and F1.

\vskip5pt \noindent
\textbf{Node Classification.}
Predicting the label of a node $u$ at time $t$ follows the same practice of a standard GCN: The activation function of the last graph convolution layer is the softmax, so that $h_t^u$ is a probability vector.

Publicly available data sets for node classification in the dynamic setting are rare. We use only one data set (Elliptic) for demonstration. This data set is the largest one in node count in Table~\ref{tab:dataset}. The evaluation metrics are the same as those for edge classification.

\subsection{Compared Methods}
We compare the two versions of the proposed method, EvolveGCN-H and EvolveGCN-O, with the following four baselines (two supervised and two unsupervised).

\vskip5pt \noindent
\textbf{GCN.} The first one is GCN without any temporal modeling. We use one single GCN model for all time steps and the loss is accumulated along the time axis.

\vskip5pt \noindent
\textbf{GCN-GRU.} The second one is also a single GCN model, but it is co-trained with a recurrent model (GRU) on node embeddings. We call this approach GCN-GRU, which is conceptually the same as Method 1 of~\cite{Seo2016}, except that their GNN is the ChebNet~\cite{Defferrard2016} and their recurrent model is the LSTM.

\vskip5pt \noindent
\textbf{DynGEM.} \cite{Goyal2017} The third one is an unsupervised node embedding approach, based on the use of graph autoencoders. The autoencoder parameters learned at the past time step is used to initialize the ones of the current time for faster learning.

\vskip5pt \noindent
\textbf{dyngraph2vec.} \cite{Goyal2019}
This method is also unsupervised. It has several variants: dyngraph2vecAE, dyngraph2vecRNN, and dyngraph2vecAERNN. The first one is similar to DynGEM, but additionally incorporates the past node information for autoencoding. The others use RNN to maintain the past node information.

\begin{table*}[ht]
\centering
\small
\caption{Performance of link prediction. Each column is one data set.}
\label{tab:link_prediction}
\begin{tabular}{|c|ccccc|ccccc|}
\hline
& \multicolumn{5}{|c|}{mean average precision} & \multicolumn{5}{c|}{mean reciprocal rank} \\
&  SBM & BC-OTC & BC-Alpha & UCI & AS & SBM & BC-OTC & BC-Alpha & UCI & AS \\
\hline
\hline
GCN               & 0.1987 & 0.0003 & 0.0003 & 0.0251 & 0.0003 & 0.0138 & 0.0025 & 0.0031 & 0.1141 & 0.0555 \\
GCN-GRU           & 0.1898 & 0.0001 & 0.0001 & 0.0114 & 0.0713 & 0.0119 & 0.0003 & 0.0004 & 0.0985 & 0.3388 \\
DynGEM            & 0.1680 & 0.0134 & 0.0525 & 0.0209 & 0.0529 & 0.0139 & 0.0921 & 0.1287 & 0.1055 & 0.1028 \\
dyngraph2vecAE    & 0.0983 & 0.0090 & 0.0507 & 0.0044 & 0.0331 & 0.0079 & 0.0916 & 0.1478 & 0.0540 & 0.0698 \\
dyngraph2vecAERNN & 0.1593 & {\bf0.0220} & {\bf0.1100} & 0.0205 & 0.0711 & 0.0120 & {\bf0.1268} & {\bf0.1945} & 0.0713 & 0.0493 \\
EvolveGCN-H       & 0.1947 & 0.0026 & 0.0049 & 0.0126 & {\bf0.1534} & {\bf0.0141} & 0.0690 & 0.1104 & 0.0899 & {\bf0.3632} \\
EvolveGCN-O       & {\bf0.1989} & 0.0028 & 0.0036 & {\bf0.0270} & 0.1139 & 0.0138 & 0.0968 & 0.1185 & {\bf0.1379} & 0.2746 \\
\hline
\end{tabular}
\end{table*}

\subsection{Additional Details}
The data set Elliptic is equipped with handcrafted node features; and Reddit contains computed feature vectors. For all other data sets, we use one-hot node-degree as the input feature. Following convention, GCN has two layers and MLP has one layer. The embedding size of both GCN layers is set the same, to reduce the effort of hyperparameter tuning.
The time window for sequence learning is 10 time steps, except for SBM and Elliptic, where it is 5.

\subsection{Results for Link Prediction}

The MAP and MRR are reported in Table~\ref{tab:link_prediction}. At least one version of EvolveGCN achieves the best result for each of the data sets SBM, UCI, and AS. For BC-OTC and BC-Alpha, EvolveGCN also outperforms the two GCN related baselines, but it is inferior to DynGEM and dyngraph2vec. These latter methods differ from others in that node embeddings are obtained in an unsupervised manner. It is surprising that unsupervised approaches are particularly good on certain data sets, given that the link prediction model is trained separately from graph autoencoding. In such a case, graph convolution does not seem to be sufficiently powerful in capturing the intrinsic similarity of the nodes, rendering a much inferior starting point for dynamic models to catch up. Although EvolveGCN improves over GCN substantially, it still does not reach the bar set by graph autoencoding.

\subsection{Results for Edge Classification}

\begin{figure}[ht]
\centering
\includegraphics[width=\linewidth]{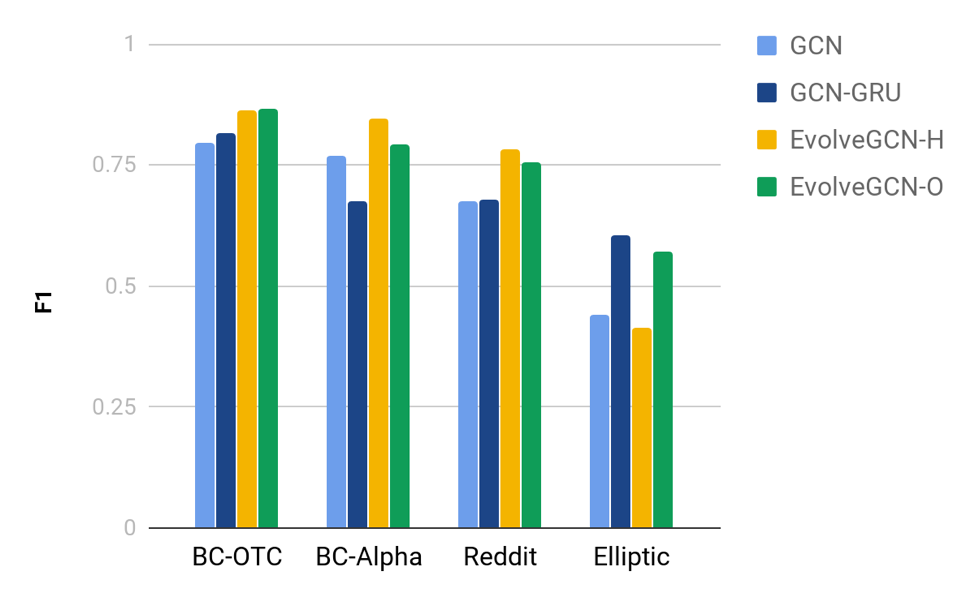}
\caption{Performance of edge classification and node classification. For edge classification (BC-OTC, BC-Alpha, and Reddit), the F1 score is the micro average. For node classification (Elliptic), because of the exceedingly high class imbalance and strong interest in the minority class (illicit transactions), the minority F1 is plotted instead.}
\label{fig:edge_classification}
\end{figure}

The F1 scores across different methods are compared in Figure~\ref{fig:edge_classification}, for the data sets BC-OTC, BC-Alpha, and Reddit. In all cases, the two EvolveGCN versions outperform GCN and GCN-GRU. Moreover, similar observations are made for the precision and the recall, which are omitted due to space limitation. These appealing results corroborate the effectiveness of the proposed method.

\subsection{Results for Node Classification}

The F1 scores for the data set Elliptic are plotted also in Figure~\ref{fig:edge_classification}. In this data set, the classes correspond to licit and illicit transactions respectively and they are highly skewed. For financial crime forensic, the illicit class (minority) is the main interest. Hence, we plot the minority F1. The micro averages are all higher than 0.95 and not as informative. One sees that EvolveGCN-O performs better than the static GCN, but not so much as GCN-GRU. Indeed, dynamic models are more effective.

\begin{figure}[!h]
\centering
\includegraphics[width=\linewidth]{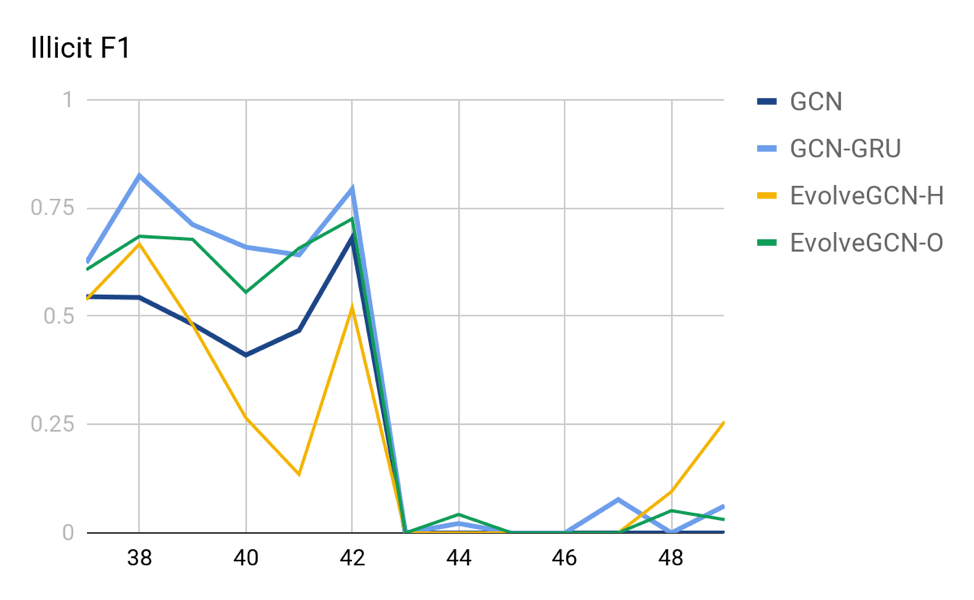}
\caption{Performance of node classification over time. The F1 score is for the minority (illicit) class.}
\label{fig:elliptic_time}
\end{figure}

For an interesting phenomenon, we plot the history of the F1 scores along time in Figure~\ref{fig:elliptic_time}. All methods perform poorly starting at step 43. This time is when the dark market shutdown occurred. Such an emerging event causes performance degrade for all methods, with non-dynamic models suffering the most. Even dynamic models are not able to perform reliably, because the emerging event has not been learned.

\section{Conclusions}
A plethora of neural network architectures were proposed recently for graph structured data and their effectiveness have been widely confirmed. In practical scenarios, however, we are often faced with graphs that are constantly evolving, rather than being conveniently static for a once-for-all investigation. The question is how neural networks handle such a dynamism. Combining GNN with RNN is a natural idea. Typical approaches use the GNN as a feature extractor and use an RNN to learn the dynamics from the extracted node features. We instead use the RNN to evolve the GNN, so that the dynamism is captured in the evolving network parameters. One advantage is that it handles more flexibly dynamic data, because a node does not need to be present all time around. Experimental results confirm that the proposed approach generally outperforms related ones for a variety of tasks, including link prediction, edge classification, and node classification.


\bibliographystyle{aaai}
\bibliography{reference}

\begin{thebibliography}{}

\bibitem[\protect\citeauthoryear{Belkin and Niyogi}{2002}]{Belkin2002}
Belkin, M., and Niyogi, P.
\newblock 2002.
\newblock Laplacian eigenmaps and spectral techniques for embedding and
  clustering.
\newblock In {\em NIPS}.

\bibitem[\protect\citeauthoryear{Bruna \bgroup et al\mbox.\egroup
  }{2014}]{Bruna2014}
Bruna, J.; Zaremba, W.; Szlam, A.; and LeCun, Y.
\newblock 2014.
\newblock Spectral networks and locally connected networks on graphs.
\newblock In {\em ICLR}.

\bibitem[\protect\citeauthoryear{Cangea \bgroup et al\mbox.\egroup
  }{2018}]{Cangea2018}
Cangea, C.; Veli\v{c}kovi\'{c}, P.; Jovanovi\'{c}, N.; and Thomas~Kipf, P.~L.
\newblock 2018.
\newblock Towards sparse hierarchical graph classifiers.
\newblock In {\em NIPS Workshop on Relational Representation Learning}.

\bibitem[\protect\citeauthoryear{Cao, Lu, and Xu}{2015}]{Cao2015}
Cao, S.; Lu, W.; and Xu, Q.
\newblock 2015.
\newblock {GraRep}: Learning graph representations with global structural
  information.
\newblock In {\em CIKM}.

\bibitem[\protect\citeauthoryear{Chen, Ma, and Xiao}{2018}]{Chen2018}
Chen, J.; Ma, T.; and Xiao, C.
\newblock 2018.
\newblock {FastGCN}: Fast learning with graph convolutional networks via
  importance sampling.
\newblock In {\em ICLR}.

\bibitem[\protect\citeauthoryear{Defferrard, Bresson, and
  Vandergheynst}{2016}]{Defferrard2016}
Defferrard, M.; Bresson, X.; and Vandergheynst, P.
\newblock 2016.
\newblock Convolutional neural networks on graphs with fast localized spectral
  filtering.
\newblock In {\em NIPS}.

\bibitem[\protect\citeauthoryear{Duvenaud \bgroup et al\mbox.\egroup
  }{2015}]{Duvenaud2015}
Duvenaud, D.; Maclaurin, D.; Aguilera-Iparraguirre, J.; G\'{o}mez-Bombarelli,
  R.; Hirzel, T.; Aspuru-Guzik, A.; and Adams, R.~P.
\newblock 2015.
\newblock Convolutional networks on graphs for learning molecular fingerprints.
\newblock In {\em NIPS}.

\bibitem[\protect\citeauthoryear{Gao and Ji}{2019}]{Gao2019}
Gao, H., and Ji, S.
\newblock 2019.
\newblock Graph {U-Nets}.
\newblock In {\em ICML}.

\bibitem[\protect\citeauthoryear{Gilmer \bgroup et al\mbox.\egroup
  }{2017}]{Gilmer2017}
Gilmer, J.; Schoenholz, S.~S.; Riley, P.~F.; Vinyals, O.; and Dahl, G.~E.
\newblock 2017.
\newblock Neural message passing for quantum chemistry.
\newblock In {\em ICML}.

\bibitem[\protect\citeauthoryear{Goyal \bgroup et al\mbox.\egroup
  }{2017}]{Goyal2017}
Goyal, P.; Kamra, N.; He, X.; and Liu, Y.
\newblock 2017.
\newblock {DynGEM}: Deep embedding method for dynamic graphs.
\newblock In {\em IJCAI Workshop on Representation Learning for Graphs}.

\bibitem[\protect\citeauthoryear{Goyal, Chhetri, and Canedo}{2019}]{Goyal2019}
Goyal, P.; Chhetri, S.~R.; and Canedo, A.
\newblock 2019.
\newblock dyngraph2vec: Capturing network dynamics using dynamic graph
  representation learning.
\newblock {\em Knowledge-Based Systems}.

\bibitem[\protect\citeauthoryear{Grover and Leskovec}{2016}]{Grover2016}
Grover, A., and Leskovec, J.
\newblock 2016.
\newblock node2vec: Scalable feature learning for networks.
\newblock In {\em KDD}.

\bibitem[\protect\citeauthoryear{Hamilton, Ying, and
  Leskovec}{2017}]{Hamilton2017}
Hamilton, W.~L.; Ying, R.; and Leskovec, J.
\newblock 2017.
\newblock Inductive representation learning on large graphs.
\newblock In {\em NIPS}.

\bibitem[\protect\citeauthoryear{Jin \bgroup et al\mbox.\egroup
  }{2017}]{Jin2017}
Jin, W.; Coley, C.~W.; Barzilay, R.; and Jaakkola, T.
\newblock 2017.
\newblock Predicting organic reaction outcomes with {Weisfeiler-Lehman}
  network.
\newblock In {\em NIPS}.

\bibitem[\protect\citeauthoryear{Kipf and Welling}{2017}]{Kipf2017}
Kipf, T.~N., and Welling, M.
\newblock 2017.
\newblock Semi-supervised classification with graph convolutional networks.
\newblock In {\em ICLR}.

\bibitem[\protect\citeauthoryear{Li \bgroup et al\mbox.\egroup }{2016}]{Li2016}
Li, Y.; Tarlow, D.; Brockschmidt, M.; and Zemel, R.
\newblock 2016.
\newblock Gated graph sequence neural networks.
\newblock In {\em ICLR}.

\bibitem[\protect\citeauthoryear{Li \bgroup et al\mbox.\egroup }{2017}]{Li2017}
Li, J.; Dani, H.; Hu, X.; Tang, J.; Chang, Y.; and Liu, H.
\newblock 2017.
\newblock Attributed network embedding for learning in a dynamic environment.
\newblock In {\em CIKM}.

\bibitem[\protect\citeauthoryear{Manessia, Rozza, and
  Manzo}{2017}]{Manessia2017}
Manessia, F.; Rozza, A.; and Manzo, M.
\newblock 2017.
\newblock Dynamic graph convolutional networks.
\newblock arXiv:1704.06199.

\bibitem[\protect\citeauthoryear{Narayan and Roe}{2018}]{Narayan2018}
Narayan, A., and Roe, P. H.~O.
\newblock 2018.
\newblock Learning graph dynamics using deep neural networks.
\newblock {\em IFAC-PapersOnLine} 51(2):433--438.

\bibitem[\protect\citeauthoryear{Nguyen \bgroup et al\mbox.\egroup
  }{2018}]{Nguyen2018}
Nguyen, G.~H.; Lee, J.~B.; Rossi, R.~A.; Ahmed, N.~K.; Koh, E.; and Kim, S.
\newblock 2018.
\newblock Continuous-time dynamic network embeddings.
\newblock In {\em WWW}.

\bibitem[\protect\citeauthoryear{Ou \bgroup et al\mbox.\egroup }{2016}]{Ou2016}
Ou, M.; Cui, P.; Pei, J.; Zhang, Z.; and Zhu, W.
\newblock 2016.
\newblock Asymmetric transitivity preserving graph embedding.
\newblock In {\em KDD}.

\bibitem[\protect\citeauthoryear{Perozzi, Al-Rfou, and
  Skiena}{2014}]{Perozzi2014}
Perozzi, B.; Al-Rfou, R.; and Skiena, S.
\newblock 2014.
\newblock {DeepWalk}: Online learning of social representations.
\newblock In {\em KDD}.

\bibitem[\protect\citeauthoryear{Roweis and Saul}{2000}]{Roweis2000}
Roweis, S.~T., and Saul, L.~K.
\newblock 2000.
\newblock Nonlinear dimensionality reduction by locally linear embedding.
\newblock {\em Science} 290(5500):2323--2326.

\bibitem[\protect\citeauthoryear{Seo \bgroup et al\mbox.\egroup
  }{2016}]{Seo2016}
Seo, Y.; Defferrard, M.; Vandergheynst, P.; and Bresson, X.
\newblock 2016.
\newblock Structured sequence modeling with graph convolutional recurrent
  networks.
\newblock arXiv:1612.07659.

\bibitem[\protect\citeauthoryear{Tang \bgroup et al\mbox.\egroup
  }{2015}]{Tang2015}
Tang, J.; Qu, M.; Wang, M.; Zhang, M.; Yan, J.; and Mei, Q.
\newblock 2015.
\newblock {LINE}: Large-scale information network embedding.
\newblock In {\em WWW}.

\bibitem[\protect\citeauthoryear{Trivedi \bgroup et al\mbox.\egroup
  }{2017}]{Trivedi2017}
Trivedi, R.; Dai, H.; Wang, Y.; and Song, L.
\newblock 2017.
\newblock {Know-Evolve}: Deep temporal reasoning for dynamic knowledge graphs.
\newblock In {\em ICML}.

\bibitem[\protect\citeauthoryear{Trivedi \bgroup et al\mbox.\egroup
  }{2018}]{Trivedi2018}
Trivedi, R.; Farajtabar, M.; Biswal, P.; and Zha, H.
\newblock 2018.
\newblock Representation learning over dynamic graphs.
\newblock arXiv:1803.04051.

\bibitem[\protect\citeauthoryear{Veli\u{c}kovi\'{c} \bgroup et al\mbox.\egroup
  }{2018}]{Velickovic2018}
Veli\u{c}kovi\'{c}, P.; Cucurull, G.; Casanova, A.; Romero, A.; Li\`{o}, P.;
  and Bengio, Y.
\newblock 2018.
\newblock Graph attention networks.
\newblock In {\em ICLR}.

\bibitem[\protect\citeauthoryear{Yu \bgroup et al\mbox.\egroup
  }{2018}]{Yu2018a}
Yu, W.; Cheng, W.; Aggarwal, C.; Zhang, K.; Chen, H.; and Wang, W.
\newblock 2018.
\newblock {NetWalk}: A flexible deep embedding approach for anomaly detection
  in dynamic networks.
\newblock In {\em KDD}.

\bibitem[\protect\citeauthoryear{Yu, Yin, and Zhu}{2018}]{Yu2018}
Yu, B.; Yin, H.; and Zhu, Z.
\newblock 2018.
\newblock Spatio-temporal graph convolutional networks: A deep learning
  framework for traffic forecasting.
\newblock In {\em IJCAI}.

\bibitem[\protect\citeauthoryear{Zhou \bgroup et al\mbox.\egroup
  }{2018}]{Zhou2018}
Zhou, L.; Yang, Y.; Ren, X.; Wu, F.; and Zhuang, Y.
\newblock 2018.
\newblock Dynamic network embedding by modeling triadic closure process.
\newblock In {\em AAAI}.

\bibitem[\protect\citeauthoryear{Zuo \bgroup et al\mbox.\egroup
  }{2018}]{Zuo2018}
Zuo, Y.; Liu, G.; Lin, H.; Guo, J.; Hu, X.; and Wu, J.
\newblock 2018.
\newblock Embedding temporal network via neighborhood formation.
\newblock In {\em KDD}.

\end{thebibliography}

\end{document}